%% file: sample_FG2021.tex
\documentclass[a4paper, 10pt, conference]{ieeeconf}      
\usepackage{FG2021}
\usepackage{graphicx}
\usepackage{subfigure}

\usepackage{hyperref}

\usepackage[accsupp]{axessibility} 

\input{math_commands}

\input{notation}

\FGfinalcopy 

\IEEEoverridecommandlockouts                              
\def\FGPaperID{38} 

\title{\LARGE \bf
Leveraging Affect Transfer Learning for Behavior Prediction in an Intelligent Tutoring System
}

\author{\parbox{16cm}{\centering
    {\large Nataniel Ruiz$^1$, Hao Yu$^1$, Danielle A.\ Allessio$^2$, Mona Jalal$^1$, Ajjen Joshi$^3$, Thomas Murray$^2$, John J.\ Magee$^4$, Jacob R.\ Whitehill$^5$, Vitaly Ablavsky$^6$, Ivon Arroyo$^5$, Beverly P.\ Woolf$^2$, \\ Stan Sclaroff$^1$, and Margrit Betke$^1$}\\
    {\normalsize
    $^1$ Boston University ~
    $^2$ University of Massachusetts Amherst ~
    $^3$ Affectiva \\
    $^4$ Clark University ~
    $^5$ Worcester Polytechnic Institute ~
    $^6$ University of Washington \\ 
    }}
}

\begin{document}

%
%
%




\IEEEoverridecommandlockouts\pubid{\makebox[\columnwidth]{978-1-6654-3176-7/21/\$31.00~\copyright{}2021 IEEE \hfill}
\hspace{\columnsep}\makebox[\columnwidth]{ }}

\ifFGfinal
\thispagestyle{empty}
\pagestyle{empty}
\else
\author{Anonymous FG2021 submission\\ Paper ID \FGPaperID \\}
\pagestyle{plain}
\fi
\maketitle

\begin{abstract}

In this work, we propose a video-based transfer learning approach for predicting problem outcomes of students working with an intelligent tutoring system (ITS). By analyzing a student's face and gestures, our method predicts the outcome of a student answering a problem in an ITS from a video feed. 
Our work is motivated by the reasoning that the ability to predict such outcomes enables tutoring systems to adjust interventions, such as hints and encouragement, and to ultimately yield improved student learning. 
We collected a large labeled dataset of student interactions with an intelligent online math tutor consisting of 68 sessions, where 54 individual students solved 2,749 problems. 
The dataset is public and available at \url{https://www.cs.bu.edu/faculty/betke/research/learning/}.
Working with this dataset, our transfer-learning challenge was to design a representation in the source domain of pictures obtained ``in the wild''  for the task of facial expression analysis, and transferring this learned representation to the task of human behavior prediction in the domain of webcam videos of students in a classroom  environment.  We developed a novel facial affect representation and a user-personalized training scheme that unlocks the potential of this representation. We designed several variants of a recurrent neural network that models the temporal structure of video sequences of students solving math problems.
Our final model, named ATL-BP for {\em Affect Transfer Learning for Behavior Prediction}, achieves a relative increase in mean F-score of 50\% over the state-of-the-art method on this new dataset.

\end{abstract}

\section{Introduction}

Research on developing intelligent tutoring systems (ITS) is a promising avenue for improving learning and education~\cite{mathspring,arroyo2014multimedia,woolf2010effect}. Previous work has shown that real-time signals from students can be used to improve their learning~\cite{arroyo2009emotion, d2010time, gordon2016affective}. Predicting whether students are having trouble with problems can allow an ITS to provide interventions, such as providing hints or encouragement, which could help the students understand or solve the problem, thus improving learning outcomes.

\begin{figure}[thpb]
    \centering
    \includegraphics[width=0.85\columnwidth]{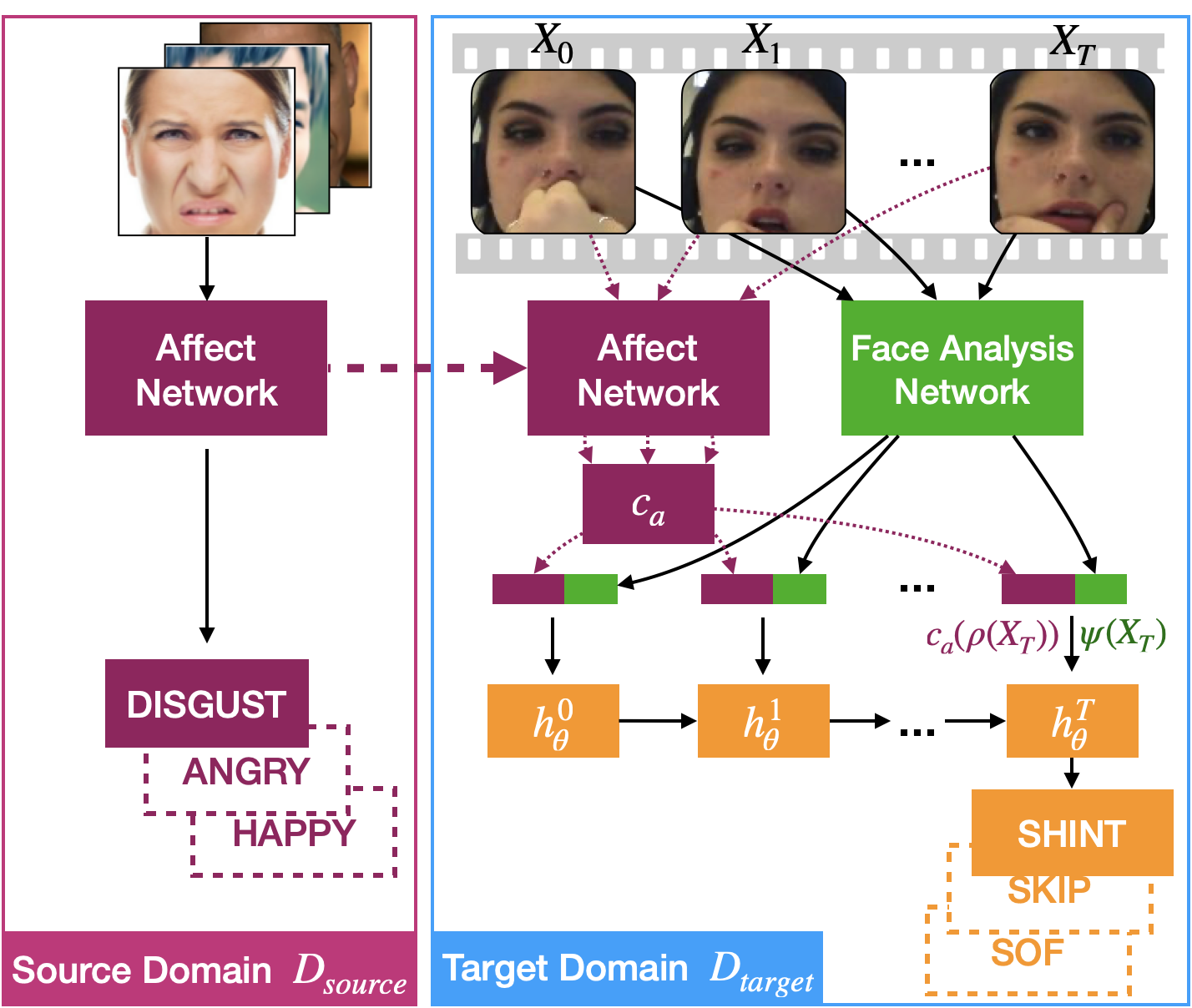}
    \caption{Our proposed {\em Affect Transfer Learning for Behavior Prediction} (ATL-BP) model for predicting the behavior of students working with an intelligent tutoring system.
    The target-domain ATL-BP model consists of three components, an affect network trained for the source domain problem of affect recognition, a facial analysis network, and an LSTM.}
    \label{fig:ATL-BP-architecture}
    \vspace{-15px}
\end{figure}

MathSpring~\cite{mathspring} is a popular online browser-based ITS that uses multimedia to encourage and support students as they solve math problems. Using the MathSpring ITS, a dataset named MathSpringSP~\cite{JoshiBeSc19} was collected, which includes 1,596 segmented videos of study sessions of students interacting with the ITS. Each problem tackled by a student has an associated outcome label automatically annotated by the ITS. Some example labels are \textit{skipped}, \textit{solved on first try}, \textit{solved with hint}, among others. In this work we address the problem of predicting the outcome label from a video feed of the student while they are solving the problem. Being able to have a model that can successfully predict outcomes while a student is completing a problem can help the ITS provide interventions such as hints or encouragement when the student is having difficulties. 

Facial and gesture analysis are valuable tools for predicting emotions, but the question of how to use them for predicting student performance with an ITS remains challenging since cues can be very subtle or ambiguous. A smile, for example, does not necessarily mean that the student is happily solving an exercise. Instead, it could indicate a student's embarrassment for not knowing the answer to a question. Moreover, in our experience, trying to obtain valid ground truth labels of the student videos from human annotators is a futile experimental task because humans have a very low accuracy rate when predicting problem outcomes from video. Just like automated facial analysis tools, human annotators struggle with interpreting the ambiguity in and limited amount of information given by student gestures.

Prior research in transfer learning for facial analysis tasks mostly focuses on transfer learning for the same task in order to bridge domain gaps such as personalization of a prediction system to specific individuals~\cite{almaev2015learning,chen2012person,chen2013learning,sangineto2014we,zen2016learning}, improving results on a benchmark by fine-tuning neural networks that are pre-trained on external datasets for a similar prediction task~\cite{kaya2017video}, or improving results by pre-training on a related facial analysis task~\cite{xu2015facial}. In contrast, our work tackles the more challenging transfer learning across domains and tasks, which is a form of \textit{transductive transfer learning}~\cite{pan2009survey},
Specifically, we tackle the problem of learning a representation in the source domain of in-the-wild pictures for the task of facial expression analysis and transferring this learned representation to the task of human behavior prediction in the domain of webcam videos in a controlled environment (Fig.~\ref{fig:ATL-BP-architecture}). While prior work has explored transfer learning from facial analysis to behavior analysis, for example, using VGG-Face facial recognition embeddings to predict driver distraction~\cite{dua2019autorate}, our work is, to the best of our knowledge, the first to propose leveraging an affect representation, learned using a deep neural network, for a behavior prediction task. Our learned affect representation is general and can be used not only for predicting problem outcomes on an ITS, but in any human behavior prediction problem where affect and expression are important cues.

The largest obstacle in training an end-to-end deep learning model for behavior analysis problems is the fact that data are relatively scarce, which increases the risk of overfitting. As a first step to alleviating the data problem, we present MathSpringSP+, an extended version of the MathSpringSP dataset, which is roughly double the size of the original dataset.  Next, we propose a novel facial affect representation for behavior prediction problems that is learned from a large affect classification dataset. We show that, by incorporating this affect embedding, we can obtain improvements compared to more traditional deep face embeddings such as the VGG-Face facial recognition embedding~\cite{BMVC2015_41}. We developed a two-layer {\em Long Short Term Memory} (LSTM) model~\cite{HochreiterSc97} that takes into account the temporal structure of the problem and successfully leverages our affect embedding. We show that, by conducting user-personalized training where a small portion of a student's initial captured data is used to fine-tune the model, our method outperforms the previous state-of-the-art method~\cite{JoshiBeSc19} by 50\%. Finally, we present a video dataset of problem-solving interactions of children and show that finetuning the ATL-BP affect network using children face images further improves the performance.
We summarize our contributions as follows:
\begin{itemize}
    \item We present MathSpringSP+: a large labeled dataset of student interactions with an intelligent online math tutor consisting of 68 sessions, where 54 individual students solved 2,749 problems. This dataset includes two views of students solving each problem as well as \textit{problem outcome labels} that describe the performance of the students on each individual problem. We will release this dataset upon acceptance.
    \item We present a transfer learning facial affect representation that can be used for behavior prediction tasks. This representation is learned from a large facial affect dataset.
    \item We are the first to model the temporal structure of video sequences of students solving math problems using a recurrent neural network architecture, improving performance on existing datasets.
    \item Our proposed {\em Affect Transfer Learning for Behavior Prediction} (ATL-BP) model outperforms the previous state-of-the-art method by 50\%.
    \item We present a dataset of children problem-solving interactions collected in the same manner as MathSpringSP+ and show that finetuning the ATL-BP affect network using children face images further improves the performance on this target domain.
\end{itemize}

\section{Related Work}

\paragraph{\textbf{Intelligent Tutoring Systems}} Intelligent tutoring systems have been evaluated and shown to produce learning gains~\cite{DMelloOlWiHa12,mayo2001optimising,mitrovic2020effect,mondragon2016evaluating,Vanlehn11}. One meta-analysis shows test score improvements from the 50th to 75th percentile~\cite{KulikFl16}. Some ITS have been shown to match the success of one-on-one human tutoring and students using these tutors outperform students from conventional classes in 92\% of the controlled evaluations and perform twice as high as for students using typical (non-intelligent systems)~\cite{corbett1992student,graesser2001intelligent,kulikjames}. 

There is a large amount of work that analyzes user affect, emotions and expressions from interactions with games or intelligent tutoring systems~\cite{amershi2016using,baker2010better,craig2004affect,DMelloOlWiHa12,d2007toward,jaques2014predicting,kapoor2007automatic,lalle2018prediction,nkambou2006framework,NyeKaToCoStAuGe18,Picard10,picard_toward_machine_emotional,woolf2009affect}. 
In certain cases the predicted affect information is used to improve learning. For example, Strain and D'Mello~\cite{StrainDm11} have studied the role of emotion in ITS engagement, task persistence, and learning gain. Gaze prediction has also been used in an effort to respond to students' boredom and to perform interventions~\cite{DMelloOlWiHa12}. Further, relationships between visual facial Action Unit (AU) factors and self-reported traits such as academic effort, study habits, and interest in the subject have been studied~\cite{NyeKaToCoStAuGe18}.

In contrast to this body of work, our work focuses on using predicted deep affect embeddings that are learned from a large facial affect dataset to improve behavior prediction in an ITS. Behavior prediction can be useful in improving learning by tailoring the interventions of the ITS to the predicted actions of the student. To the best of our knowledge, our work is the first to use an affect embedding for behavior prediction in an ITS.

\paragraph{\textbf{Interventions in an Online Tutor}}
Prior research has examined the impact of several interventions in ITS to improve student outcome and affect, specifically, affective messages delivered by avatars and empathetic messages that responded to students' recent emotions~\cite{woolf2010effect}. Interventions in the MathSpring ITS led to improved grades in state standardized exams~\cite{CraigGrPe18} as well as influence students' perceptions of themselves as learners~\cite{KarumbaiahLiAlWoArWi17}. Empathetic characters which provide interventions generate superior results both to improve student interactions with the system, address negative student emotions, and in the overall learning experience~\cite{Kim05}. Predicting outcomes of problems for students is a valuable source of information for planning and executing ITS interventions for improving learning~\cite{Delgado-etal-2021,Yu-etal-2021}. For example the ITS could provide hints when the system predicts that the student will not be able to successfully complete the problem.

\paragraph{\textbf{Predicting Exercise Outcome}}
Joshi et al.~\cite{JoshiBeSc19} presented a first attempt at tackling the problem of exercise outcome prediction. They did not explore deep learning representations but used traditional facial analysis features such as head pose, gaze and facial action units (AUs). They also did not attempt to model the temporal component of the videos, which is a rich source of information, and instead opted to summarize features from a video into one single feature vector.  The method by Joshi et al.~\cite{JoshiBeSc19} can be considered the previous state of the art in student outcome prediction, and, thus, our experimental results include a performance comparison between this method and our models.

\section{MathSpringSP+ Dataset}

In order to build an ITS capable of understanding student behavior and producing interventions, it is critical to build tailored datasets that allow development of behavior understanding techniques. To this end, in this work we expand the MathSpringSP dataset described by Joshi et al.~\cite{JoshiBeSc19}, following the same data collection protocol. We name the extended dataset MathSpringSP+. MathSpringSP+ is roughly double the size of the original MathSpringSP dataset.

\begin{figure*}  
\centering
\includegraphics[width=0.7\textwidth]{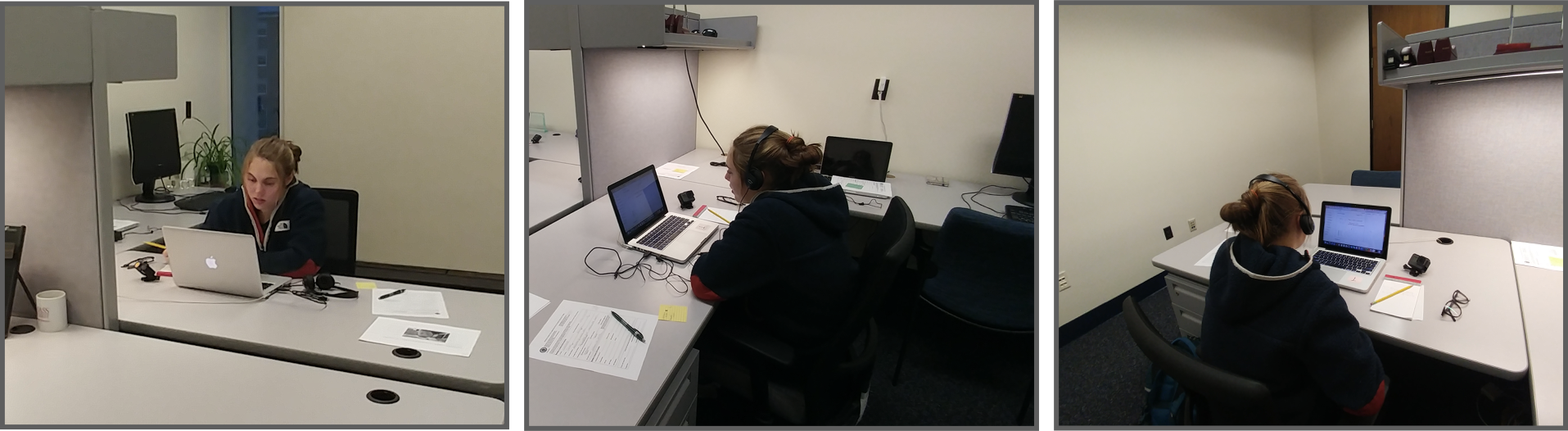}
\vspace{-5px}
\caption[]{Data capture setup for the MathSpringSP+ dataset from three views (front, side and back). The student completes problems on a laptop. The laptop webcam and Go-Pro camera on the right side of the student are used to capture the student's upper body and face during the completion of problems.}
\label{fig:datacapturesetup}
\vspace{0px}
\end{figure*}

MathSpringSP+ consists of Webcam and GoPro videos that are recorded while college students solve math problems using the online tutor MathSpring~\cite{mathspring} on a laptop. The webcam is positioned on the laptop and films the student at a frontal angle. The designated spot for the GoPro camera is above the notepad, to the front-right of the student. Figure~\ref{fig:datacapturesetup} illustrates our data capturing setup from three viewpoints. Students work on solving math problems for 30--40 minutes or approximately 50 problems. The number of problems solved is variable between sessions depending on the rate at which each student solves problems. We divide each student's video session into shorter video segments, where each segment is associated with an individual math problem. Each math problem video clip has an associated problem outcome $y$, recorded in the log files of the ITS~\cite{JoshiBeSc19}. This problem outcome is automatically labeled by the software using a rule-based algorithm that chooses from the following seven possible student outcomes:
\begin{itemize}
    \item ATT (attempted): Student did not see any hints and solved the problem after one incorrect attempt
    \item GIVEUP: Student tried to answer the problem or asked for a hint but ultimately skipped the problem
    \item GUESS: Student did not see hints, but solved the problem after more than one incorrect attempts
    \item NOTR (not read): Student performed some action, but the first action was too fast for the student to have read the problem
    \item SHINT (solved with hint): Student eventually submitted the correct answer after seeing one or more hints
    \item SKIP: Student skipped the problem without asking for a hint or attempting to answer the problem
    \item SOF (solved on the first attempt): Student answered correctly on the first attempt, without seeing any hints
\end{itemize}

\begin{figure*}
  \centering
 \includegraphics[width=0.67\textwidth]{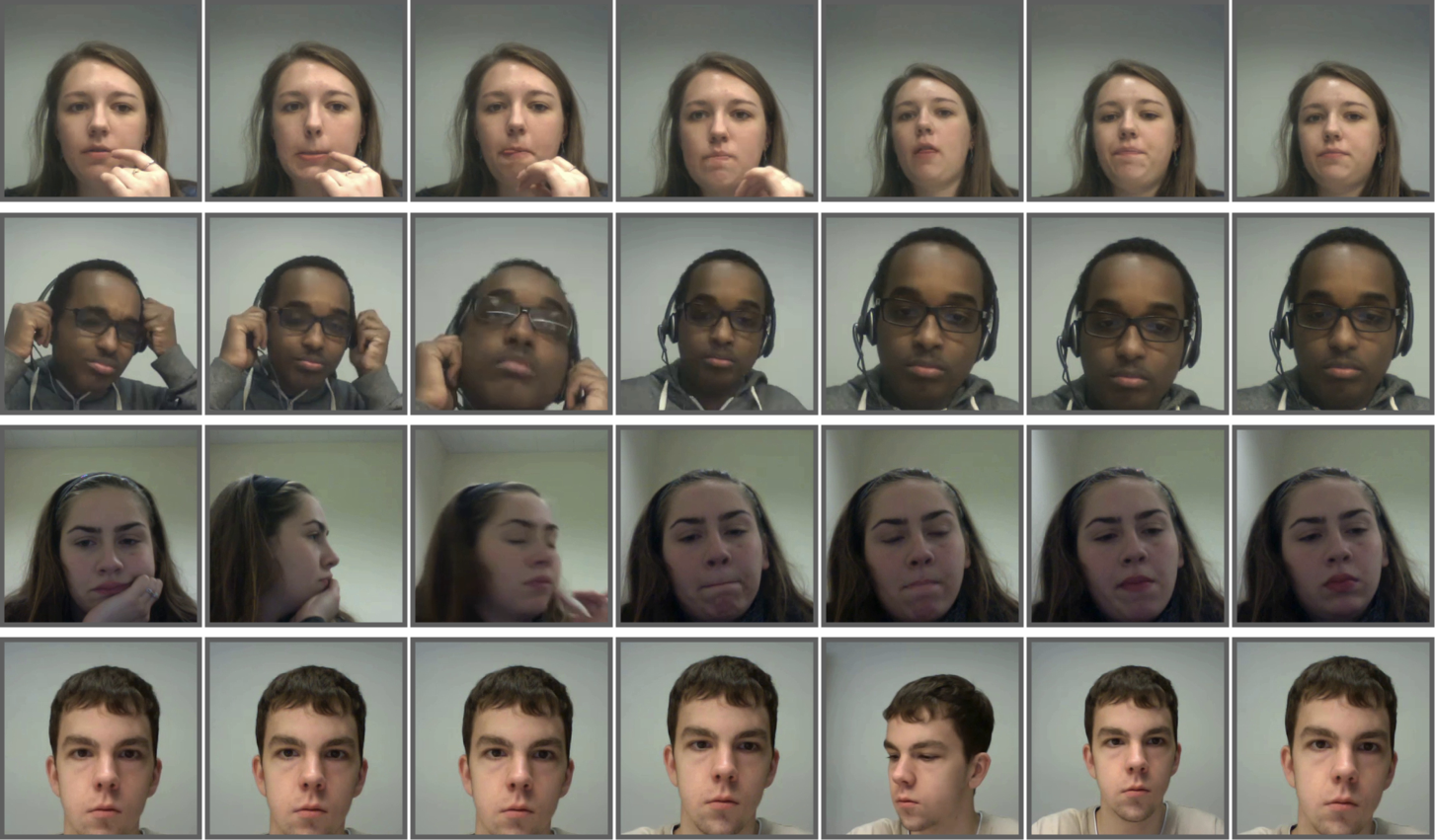}
 \vspace{-5px}
 \caption{Example face-cropped images from the MathSpringSP+ dataset showing the evolution of student expressions. In particular we notice changes in head pose, hand gestures, face occlusion and facial gestures throughout the videos. Expressions in videos can be very subtle, as well as ambiguous, making the prediction problem challenging.} 
 \label{fig:faces_01}
 \vspace{-10px}
\end{figure*}

Examples of the variation in student facial expression throughout the process of answering problems in the math tutor are shown in Figure \ref{fig:faces_01}. We note that expressions can be very subtle. Expressions can also be ambiguous: a frown can mean that the student is very focused and will solve the problem correctly or that they are having difficulties with the problem. Expression intensities and variance depend on the individual, and it is challenging to generalize to different identities. Finally, our method has to deal with hand gestures, face occlusions and extreme pose changes, some of which are shown in Figure \ref{fig:faces_01}. 
A total of 24 students participated in the extended study, compared to 30 in the original study.  The dataset will be made publicly available. We note that the dataset only includes individuals who have provided written consent that their data may be used publicly for research purposes.
Several students participated in multiple sessions. Each session lasted approximately one hour. In total, 30 student sessions were recorded, which yielded 1,153 problem samples. Thus, the extended MathSpringSP+ dataset contains videos of a total of 54 unique students, 68 student sessions and 2,749 problem samples. This amount of data almost doubles the original MathSpringSP dataset, which contains 38 student sessions and 1,596 problem samples. A detailed breakdown of the relative sizes of MathSpringSP and MathSpringSP+ are shown in Table~\ref{table:dataset}.

\begin{table}[t]
\caption{Size comparison of our extended MathSpringSP+ dataset compared to MathSpringSP}
\label{table:dataset}

  \setlength{\tabcolsep}{0.7pt} 
\renewcommand{\arraystretch}{1} 

\centering

\resizebox{0.72\columnwidth}{!}{%

\begin{tabular}{ lcc } 
 & MathSpringSP & \ \ MathSpringSP+ \\
\hline
Individual Students & 30 & \ \ 54 \\
Student Sessions & 38 & \ \ 68  \\
Problem Samples & 1,596 & \ \ 2,749 \\
\hline
\end{tabular}

}
\end{table}

\section{Method}

The dataset consists of labeled video pairs $(\video, \vlabel)$, where the video~$\video$ is a time series of RGB frames $\video = \{\video_t \ | \ t=1..T\}$ of a student solving a problem, and the scalar label $\vlabel$ indicates the outcome class for that problem.  The task at hand is a 7-label classification problem, i.e., $y \in \{1, ..., C\}$, for
$C=7$. 

Our challenge was to work out how to leverage state-of-the-art affect recognition techniques to compute an output label $y$ from the input video $\video$.  Affect recognition models provide affect estimates from images of faces that typically show strong emotions, e.g., the disgust expressed in the women's face on the left in Fig.\ref{fig:ATL-BP-architecture}. We decided to use a ResNet-50 network~\cite{he2016deep} and the AffectNet dataset~\cite{MollahosseiniHaMa17}, which contains more than one million facial images collected ``in the wild'' from the Internet, to solve the source domain problem of predicting eight emotions, neutral, happiness, sadness, surprise, fear, disgust, anger, contempt, plus the two classes, uncertain, and non-face. We employ this trained affect network to solve the target-domain problem of student outcome prediction.

The proposed ATL-BP model consists of three main components, see Figure~\ref{fig:ATL-BP-architecture}, the affect network, a facial analysis network, and an LSTM. We also study variants of our model by either removing the affect network or replacing it with a face recognition network.

First, from the last layer of the trained affect network, ATL-BP extracts a fixed-size embedding of size 8,192, computed for each frame $\video_t$, and compresses it into a lower-dimensional vector~$\rho(\video_t)$ by learning the weights for a fully-connected neural network layer~$\fclayeraffect$ (Fig.~\ref{fig:ATL-BP-architecture}, magenta).  

Second, ATL-BP uses a facial analysis model to extract facial Action Unit (AU) presence and intensity, gaze direction, and head pose for each frame~$\video_t$. We note these traditional facial analysis features as $\psi(\video_t)$ (Fig.~\ref{fig:ATL-BP-architecture}, green). We chose the OpenFace 2.0 model~\cite{BaltrusaitisZaLiMo18} to compute student head position, head pose, gaze, facial AU presence, and facial AU intensity from individual frames in each video segment.
  
For our main ATL-BP model we devised a feature representation that is based on concatenating the outputs of our proposed affect representation and the facial analysis components:
\[
\phi(\video_t) =  \fclayeraffect(\rho(\video_t)) \concat \psi(\video_t),
\]
where $\concat$ is the concatenation operation for vectors. The compressed embedding $\fclayeraffect(\rho(\video_t))$ is 100-dimensional. The full feature vector $\phi(\video_t)$ has dimension 149 for every frame $video_t$.

For our model variants, we replace the affect network by a face recognition model in order to extract face related features. We selected the pre-trained VGG-Face network~\cite{BMVC2015_41}, which computes an embedding $\xi$ of dimension 2,622. ATL-BP compresses the feature representation~$\xi(\video_t)$, computed by this network for each video frame~$\video_t$, using another fully-connected layer $\fclayervgg$, into $\fclayervgg(\xi(\video_t))$.

Finally, in order to model the temporal nature of the videos, we designed a unidirectional 2-layer LSTM classifier $\model$ with 200 hidden units that processes the feature vector $\phi(\video_t)$ frame by frame and produces the final estimate of student outcome~$y$ (Fig.~\ref{fig:ATL-BP-architecture}, orange).



\section{Experiments}

We present experiments on problem outcome prediction on the MathSpringSP+ dataset. These experiments study our contributions, which include incorporating temporal information from video streams by using an LSTM and using our affect transfer learning representation.  The experiments also show how user-personalized training unlocks the effectiveness of our affect representation. We also study early prediction as well as present ablation studies for the dimensionality reduction that is accomplished by the proposed fully-connected layer. In this work we limit ourselves to the webcam video stream of the student. 
 
\subsection{Model Training}

\paragraph{\textbf{Training the Affect Representation Network}}
For source domain affect training, we selected a ResNet-50 network. We pre-trained the affect network on a subset of 50,000 randomly sampled images from the AffectNet dataset and validated the network on 5,000 randomly selected images. We limited ourselves to a subset since the dataset contains more than one million examples.  Note that our training and validation data subsets are not the same as used by  \cite{MollahosseiniHaMa17}.  On our subset, our network achieves a mean accuracy of $47.3\%$, which is close to the accuracy reported by \cite{MollahosseiniHaMa17} on their skew-normalized validation set of $54\%$, and much higher than the random baseline of $9.0\%$.   The relatively low accuracy scores can be contributed to a data that is unbalanced, noisy, and overall challenging.

We extracted the target domain affect features from our videos by performing inference of the affect network on every frame. We chose a granularity of three frames per second, down from 30 frames per second in our videos, in order to save on processing time and storage space. We found that this granularity was a good compromise between performance and cost. The affect network uses each frame as an input and the last-layer features are extracted as a vector of dimension $8,192$.

We trained the affect network with the Adam optimizer with a learning rate of $3\times10^{-4}$, $\beta_1$ of $0.9$, and $\beta_2$ of $0.999$. The standard batch normalization layers of the ResNet-50 were used and fixed throughout training.

\paragraph{\textbf{Training ATL-BP to Predict Student Exercise Outcome}}
For each frame used, the feature vector computed is $\phi(\video_t) = \psi(\video_t) \concat \fclayeraffect(\rho(\video_t)) \concat \fclayervgg(\xi(\video_t))$. 
We observed that the dimensionality reduction due to the compression layer stabilizes training and improves performance. The feature vector $\phi$ is used to train the LSTM with two stacked layers. Specifically, at each instant $t$, features $\phi(\video_t)$ are fed to the LSTM. The LSTM is trained on all the video segments.  It outputs a class probability for each problem outcome.  The LSTM is trained using the cross-entropy loss function. The Adam optimizer is used for training. We use a learning rate of $3 \times 10^{-5}$ for $30$ epochs, and a batch size of $1$.

\subsection{Experimental Setup for Testing}

\paragraph{\textbf{Model Variations for Testing}}
In addition to our main proposed ATL-BP, shown in Figure~\ref{fig:ATL-BP-architecture} and which we call 
``ATL-BP with affect embedding'' for clarity, we implemented and test two variants of ATL-BP.  The first variant is {\em ATL-BP without transfer learning.} In this model, the LSTM directly interprets the output $\psi$ of the facial analysis network and does not use the embedding scheme we propose in this work.  The second variant is {\em ATL-BP with VGG-Face embedding.}  In this model, the LSTM interprets the output $\fclayervgg(\xi(\video_t))$ concatenated with the output $\psi$ of the facial analysis network.

Furthermore, for comparison baselines, we reproduced the method described by Joshi et al.~\cite{JoshiBeSc19} and show results for a majority vote classifier. The majority vote classifier simply selects the most prevalent class in our dataset, ``Solved on First Try,'' for every video.

\paragraph{\textbf{Random Dataset Split}}
Following the experimental setup in \cite{JoshiBeSc19}, we performed five-fold cross validation on our dataset by randomly shuffling video segments and constructing five different train and test splits. The train splits contain 80\% of the data while the test splits contain the rest. 

Experiments conducted using this random splitting experimental setup cannot reliably measure generalization to new users since videos of problems from the same student can be present in both the training and test set. This means that the network does not have to learn how to generalize to a completely new identity. We propose an improved experimental setup next.

\paragraph{\textbf{User Generalization Split}}
In order to test generalization to new users we propose a leave-users-out experimental setup where users are exclusively split into either the training or test set. In other words, we enforce the rule that no video clips of the same user can be in both the test and training sets. In this manner we can measure how the system performs when applied to an unseen user. This is a substantially more challenging task since the network has to generalize to new identities and features. We suggest that all future research on this dataset use this type of setup. We created five leave-users-out splits for five-fold cross-validation and train different model variations for each split.

\subsection{Results and Discussion}

\paragraph{\textbf{ATL-BP Results for Random Splits}}
Using the experimental protocol of a random dataset split, our ATL-BP for problem outcome prediction on MathSpringSP+ achieves an accuracy of 60.2\% (Table~\ref{table:results-random}). Compared to the previous state-of-the-art method~\cite{JoshiBeSc19}, this is an increase of 14 percent points (pp) in accuracy.  ATL-BP also achieves a 44\% relative increase in mean F-score improving from 0.238 to 0.330.  The mean F-score is computed by first computing the individual F-score for all classes and averaging over all classes.  By comparing the results for ATL-BP without transfer learning and those by Joshi et al.~\cite{JoshiBeSc19}, we can see that by integrating an LSTM architecture that allows for modeling the temporal component in the videos we can achieve a marked increase in performance (5.6 pp). We achieve a further increase in performance by using deep embeddings (8.6 pp for using the VGG-Face embedding $\xi$), and especially our proposed affect embedding $\psi$ (as mentioned, 14 pp).

\begin{table}
\caption{Results for problem outcome prediction on the MathSpringSP+ dataset using five-fold cross-validation and random data splits}
\label{table:results-random}

  \setlength{\tabcolsep}{0.7pt} 
\renewcommand{\arraystretch}{1} 

\centering

\resizebox{\columnwidth}{!}{%

\begin{tabular}{ lcc } 
Method & Mean F-Score & \ \ Accuracy \\
\hline
Majority Vote Classifier & $0.103$ & \ \ $56.1\%$ \\
Joshi et al.~\cite{JoshiBeSc19} & $0.228$ & \ \ $46.2\%$  \\
ATL-BP w/o transfer learning & $0.295$ & \ \ $51.8\%$ \\
ATL-BP w/ VGG-Face embedding & $0.304$ & \ \ $54.8\%$ \\
ATL-BP w/ affect embedding & $\bf{0.330}$ & \ \ $\bf{60.2\%}$ \\
\hline
\end{tabular}

}
\vspace{-5px}
\end{table}

\paragraph{\textbf{MathSpringSP Results}}
We conducted experiments on the original MathSpringSP dataset
in order to verify that our proposed ATL-BP model with affect embeddings achieves improved results in the same testing environment presented by Joshi et al.~\cite{JoshiBeSc19}.  Our results show a consistent improvement in mean F-score and accuracy of our method (Table \ref{table:results-original}).

\begin{table}
\caption{Results for problem outcome prediction on the original MathSpringSP for ATL-BP following the data setup from Joshi et al.~\cite{JoshiBeSc19}}
\label{table:results-original}

  \setlength{\tabcolsep}{0.7pt} 
\renewcommand{\arraystretch}{1} 

\centering

\resizebox{0.95\columnwidth}{!}{%

\begin{tabular}{ lcc } 
Method & Mean F-Score & \ \ Accuracy \\
\hline
Joshi et al.~\cite{JoshiBeSc19} & $0.270$ & \ \ $54.0\%$  \\
ATL-BP w/ affect embedding & $\bf{0.362}$ & \ \ $\bf{61.0\%}$ \\
\hline
\end{tabular}
}
\vspace{-10px}
\end{table}

\paragraph{\textbf{Early Prediction of Problem Outcome}}

We experimented with obtaining prediction using only the five first seconds of each video clip (Table \ref{table:results-random-early}).  Early outcome prediction is important since the ITS should have time to react and deliver the intervention should it be decided to do so. It turns out that is straightforward to do early prediction using an LSTM since it outputs a prediction at every time step, as opposed to the method proposed by Joshi et al.~\cite{JoshiBeSc19}, where each video has to be summarized into a fixed-sized vector before being fed into a multilayer perceptron. We observe that ATL-BP achieves a large increase (6.7 pp) in performance over \cite{JoshiBeSc19}. ATL-BP without transfer learning obtains the best F-score (0.295) in this experimental setup.

\begin{table}
\caption{Results for early prediction of problem outcome using only the first five seconds of video footage on the MathSpringSP+ dataset (five-fold cross-validation, random data splits).}
\label{table:results-random-early}

  \setlength{\tabcolsep}{0.7pt} 
\renewcommand{\arraystretch}{1} 

\centering

\resizebox{\columnwidth}{!}{%

\begin{tabular}{ lcc } 
Method & Mean F-Score & \ \ Accuracy \\
\hline
Majority Vote Classifier & $0.103$ & \ \ $56.1\%$ \\
Joshi et al.~\cite{JoshiBeSc19} & $0.173$ & \ \ $46.7\%$  \\
ATL-BP w/o transfer learning & $\textbf{0.295}$ & \ \ $51.8\%$ \\
ATL-BP w/ VGG-Face embedding & $0.239$ & \ \ $47.0\%$ \\
ATL-BP w/ affect embedding & $0.270$ & \ \ $\bf{53.4\%}$ \\
\hline
\end{tabular}

}
\vspace{-10px}
\end{table}

\paragraph{\textbf{Deep Embedding Dimensionality Reduction}}
We performed an ablation study on the fully-connected layer that is used for reducing the dimensionality of the deep embeddings that are used as inputs for our LSTM architecture (Table \ref{table:dimreduc-ablation}). While the mean F-score does not change on both the VGG-Face and proposed affect embedding ATL-BP variants, dimensionality reduction does improve the accuracy of the models by 3.5 pp and 1.5 pp, respectively.

\begin{table}
\caption{Embedding dimensionality reduction ablation study. We show results for problem outcome prediction on the MathSpringSP+ dataset using five-fold cross-validation and random data splits}
\label{table:dimreduc-ablation}

  \setlength{\tabcolsep}{0.7pt} 
\renewcommand{\arraystretch}{1} 

\centering

\resizebox{\columnwidth}{!}{%

\begin{tabular}{ lcc } 
Method & Mean F-Score & \ \ Accuracy \\
\hline
ATL-BP w/ VGG-Face & $0.304$ & \ \ $51.3\%$ \\
ATL-BP w/ VGG-Face \& dim. reduction & $0.304$ & \ \ $\textbf{54.8\%}$ \\
\hline
\hline
ATL-BP w/ affect & $0.330$ & \ \ $58.7\%$ \\
ATL-BP w/ affect \& dim. reduction & $0.330$ & \ \ $\bf{60.2\%}$ \\
\hline
\end{tabular}
}
\vspace{-5px}
\end{table}

\begin{table}
\caption{Results for problem outcome prediction on the MathSpringSP+ dataset using five-fold cross-validation and the more challenging leave-users-out splits}
\label{table:results-user}
  \setlength{\tabcolsep}{0.7pt} 
\renewcommand{\arraystretch}{1} 
\centering
\resizebox{\columnwidth}{!}{%
\begin{tabular}{ lcc } 
Method & Mean F-Score & \ \ Accuracy \\
\hline
Majority Vote Classifier & $0.102$ & \ \ $55.9\%$ \\
Joshi et al.~\cite{JoshiBeSc19} & $0.182$ & \ \ $41.9\%$  \\
ATL-BP w/o transfer learning & $\bf{0.270}$ & \ \ $50.3\%$ \\
ATL-BP w/ VGG-Face embedding & $0.246$ & \ \ $51.8\%$ \\
ATL-BP w/ affect embedding & $0.251$ & \ \ $\textbf{54.0\%}$ \\
\hline
\end{tabular}
}
\vspace{-10px}
\end{table}

\paragraph{\textbf{ATL-BP Results for User Generalization}}
For the user generalization split of the training and testing data, we report the mean F-score and mean accuracy in Table~\ref{table:results-user} for the ``Majority Vote Classifier'' benchmark, Joshi et al.~\cite{JoshiBeSc19} and our proposed model with different combinations of embeddings. 
We observe that the temporal modeling improves results from Joshi et al.~\cite{JoshiBeSc19} substantially (12.1 pp in accuracy). We observe that ATL-BP without transfer learning outperforms the ATL-BP version with our proposed affect embedding with regards to the F1 score. We hypothesize that leveraging affect embeddings is more difficult in this setup since the model does not have access to baseline levels of expression for each user.

\paragraph{\textbf{Personalization of Prediction}}
An effective real-time tutoring system would benefit from personalizing its prediction using initial data captured from a specific user stream. People have different emotional and expression baselines that can be learned using data collected in a trial run of the system. Specifically, we want the model to act on the variations of our affect embedding compared to the mean affect embedding, since each person will have a different baseline expression and thus a different baseline affect embedding. The model does not have any way to integrate this information without it being personalized for each user.

We propose a personalization scheme in which our system can be tailored to individual users and can fully utilize our proposed affect embedding. In this scheme, the network is fine-tuned on the initial problems corresponding to 20\% of the session for users in the test set for 30 epochs. Our experiments show that user personalization unlocks the potential of the affect features (Table \ref{table:results-pers}). ATL-BP with affect embedding achieves the highest F-score of 0.308 and the highest accuracy of 55.1\% compared to the other methods. Our full method achieves a relative increase of 50\% in mean F-score as well as an absolute increase in accuracy of more than 11 pp compared to the previous state of the art~\cite{JoshiBeSc19}. Our full method also outperforms variants of ATL-BP, which do not use our proposed affect representation.

\begin{table}
\caption{Results for problem outcome prediction (7-classes) on the MathSpringSP+ dataset after user personalization (five-fold cross-validation and leave-user-out splits)}
\label{table:results-pers}

  \setlength{\tabcolsep}{0.7pt} 
\renewcommand{\arraystretch}{1} 

\centering

\resizebox{\columnwidth}{!}{%

\begin{tabular}{ lcc } 
Method & Mean F-Score & \ \ Accuracy \\
\hline
Majority Vote Classifier & $0.090$ & \ \ $45.3\%$ \\
Joshi et al.~\cite{JoshiBeSc19} & $0.206$ & \ \ $43.8\%$  \\
ATL-BP w/o transfer learning & $0.278$ & \ \ $48.4\%$ \\
ATL-BP w/ VGG-Face embedding & $0.262$ & \ \ $48.7\%$ \\
ATL-BP w/ affect embedding & $\bf{0.308}$ & \ \ $\bf{55.1\%}$ \\
\hline
\end{tabular}

}
\vspace{-5px}
\end{table}

\paragraph{\textbf{Outcome Prediction for Children}}

\begin{figure}
  \centering
 \includegraphics[width=\columnwidth]{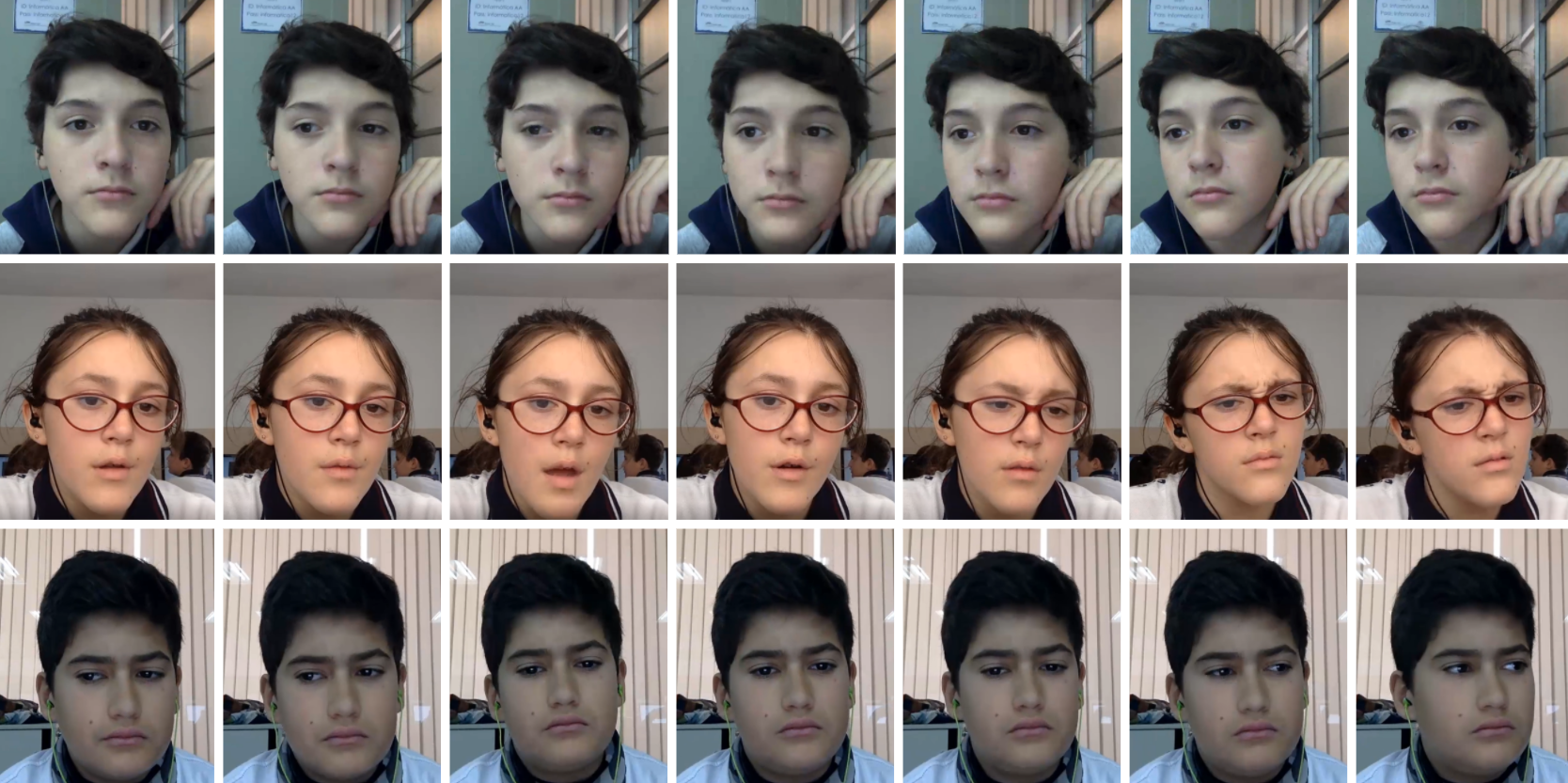}
 \vspace{-10px}
 \caption{Example face-cropped images from the Children dataset showing the evolution of student expressions.} 
 \label{fig:faces_children}
 \vspace{-15px}
\end{figure}

As a final experiment we tested our method on a new dataset of children working on math problems. Following the same data collection protocol as MathSpringSP+, we collected 968 recorded problem-solving interaction samples of fifty-one K12 students who used MathSpring. We show some extracted frames from the dataset in Figure \ref{fig:faces_children}. Results on this Children dataset show that our model consistently outperforms the baseline and previous state-of-the-art method (Table~\ref{table:results-children}).

Since the AffectNet dataset mainly captures facial expressions of adults, we further finetuned the affect representation network using two datasets of children facial expressions, LIRIS~\cite{khan2019novel} and CAFE~\cite{lobue2015child}, in order to tailor the model specifically for children. LIRIS contains 208 video clips of 6-to-12-year-old children showing six basic spontaneous facial expressions, while CAFE dataset contains 1,192 images of 2-to-8-year-old children posing for seven facial expressions. We trained three variants of models using LIRIS only (frames), CAFE only, and a combination of both datasets. The best model among the three achieves the highest accuracy (45.2\%) and mean F-score (0.278), improving on the previous state-of-the-art~\cite{JoshiBeSc19} (13.2 pp absolute increase in accuracy and 38\% relative increase in mean F-score) on the challenging task of predicting future outcome using only student face movements and gestures. The prediction task has 7 classes which contributes to the difficulty.

 \begin{table}
\caption{Results for problem outcome prediction (7-classes) on the Children dataset (five-fold cross-validation, random data splits).}
\label{table:results-children}

  \setlength{\tabcolsep}{0.7pt} 
\renewcommand{\arraystretch}{1} 

\centering

\resizebox{\columnwidth}{!}{%
\begin{tabular}{ lcc } 
Method & Mean F-Score & \ \ Accuracy \\
\hline
Majority Vote Classifier & $0.070$ & \ \ $32.3\%$ \\
Joshi et al.~\cite{JoshiBeSc19} & $0.202$ & \ \ $32.0\%$  \\
ATL-BP w/o transfer learning & $0.238$ & \ \ $33.4\%$ \\
ATL-BP w/ affect embedding & $0.260$ & \ \ $39.6\%$ \\
ATL-BP w/ LIRIS children affect embedding & $0.272$ & \ \ $45.2\%$ \\
ATL-BP w/ CAFE children affect embedding & $0.273$ & \ \ $44.4\%$ \\
ATL-BP w/ LIRIS+CAFE affect embedding & $\textbf{0.278}$ & \ \ $\textbf{45.2\%}$ \\
\hline
\end{tabular}
}
\vspace{-10px}
\end{table}

\section{Conclusion}

We introduced a large labeled dataset of student interactions with an intelligent online math tutor that consists of 68 sessions, where 54 individual students solved 2,749 math problems. Using this dataset we design a transfer learning model ATL-BP that improves problem outcome predictions for students interacting with the ITS and answering math problems. By modeling the temporal structure of the videos with ATL-BP, we achieved a substantial increase in classification F-score and accuracy compared to previous state of the art in this task. Additionally, using a novel affect representation along with user personalization, we achieved a further increase in performance. More generally, these promising results suggest that leveraging affect representations might be valuable in behavior analysis applications more generally. Our final method achieves a $50\%$ relative increase in mean F-score as well as an absolute 11 percentage point increase in accuracy compared to previous work. Finally, we collect a dataset of children student interactions and present results on this dataset. We show that finetuning of the Affect network with age-appropriate images and video further improves performance in this scenario. These results pave the way for future improvements in solutions for this task.  Future tutor systems may use our proposed outcome prediction model in order to deliver real-time interventions to improve the learning of students.

\section*{Acknowledgements}

We thank the participants of our experimental study and acknowledge partial funding for this work by the National Science Foundation, grant 1551572.

{\small
\bibliographystyle{ieee}
\bibliography{egbib}
}

\end{document}

%% file: math_commands.tex
\usepackage{amsmath,amsfonts,bm}

\def\eqref#1{equation~\ref{#1}}

\def\1{\bm{1}}

\def\vy{{\bm{y}}}

\def\vX{{\bm{X}}}

\DeclareMathAlphabet{\mathsfit}{\encodingdefault}{\sfdefault}{m}{sl}
\SetMathAlphabet{\mathsfit}{bold}{\encodingdefault}{\sfdefault}{bx}{n}



%% file: notation.tex
\def\video{\vX}
\def\vlabel{\vy}

\def\model{h_{\theta}}

\def\fclayeraffect{c_{a}}
\def\fclayervgg{c_{v}}

\def\concat{\oplus}
